\documentclass{article}
\usepackage{iclr2027_conference,times}
\usepackage[T1]{fontenc}
\usepackage{amsmath,amssymb}
\usepackage{graphicx}
\usepackage{booktabs,array,tabularx}
\usepackage{xcolor}
\usepackage{placeins}
\usepackage{hyperref}
\usepackage{url}
\hypersetup{colorlinks=true,linkcolor=black,citecolor=black,urlcolor=blue,
  pdftitle={PileBelief: Persistent Physical State for Interaction-Driven World Modeling},
  pdfauthor={Hongyi Lin, Song Zhang, Haiquan Liu, Yang Liu, Jinhua Zhao, Xiaobo Qu}}

\newcommand{\method}{PileBelief}
\newcommand{\tablesize}{\normalsize}

\title{PileBelief: Persistent Physical State for\\
Interaction-Driven World Modeling}

\author{%
\textbf{Hongyi Lin}$^{1,2}$ \enspace
\textbf{Song Zhang}$^{3}$ \enspace
\textbf{Haiquan Liu}$^{3}$ \enspace
\textbf{Yang Liu}$^{1}$ \enspace
\textbf{Jinhua Zhao}$^{2}$ \enspace
\textbf{Xiaobo Qu}$^{1}$\\[4pt]
{\normalfont\fontsize{8.5}{10}\selectfont
$^{1}$Tsinghua University \quad
$^{2}$Massachusetts Institute of Technology \quad
$^{3}$Tsing-AI(Shanghai) Technology Co., Ltd}\\[3pt]
{\normalfont\small
Email: \texttt{thu\_ets\_ly@tsinghua.edu.cn}}
}

\iclrfinalcopy

\begin{document}
\maketitle

\fancyhead{}
\fancyhead[L]{Preprint}

\begin{abstract}
World models allow robots to anticipate action consequences before execution. This capability is especially valuable in excavation, where each scoop reshapes the terrain and affects subsequent actions. Local observations, however, cannot fully reveal the underlying support and material conditions. We present PileBelief, an interaction-driven persistent world model for partially observed excavation that retains physical evidence beyond the visible surface. It combines an observation-conditioned physical prior with world-addressed deformation memory and physical-response memory. Action-aligned reads and gated residual corrections refine terrain-change and outcome predictions. With deployment weights fixed, completed interactions update measured belief, while hypothetical actions advance a separate imagined state. Compared with a current-observation-only baseline, PileBelief reduces five-step joint prediction error by 10.8\% and offline action-selection regret by 65.5\%. Experiments on Newton/MPM and real excavation datasets further demonstrate improved terrain-change and bucket-volume prediction. Our method enables multi-step prediction and candidate-action ranking from local observations, even when the underlying soil state is unknown. These results identify persistent physical belief as a useful representation for world models of environments that robots continually reshape.

\end{abstract}

\section{Introduction}
World models learn how an environment evolves under action. Their internal state connects observations to future consequences, enabling an agent to imagine trajectories and evaluate actions before execution \citep{planet,dreamer,dreamerv3}. In environments that a robot physically modifies, this state has a particularly demanding role. An action changes both what is visible and the hidden conditions governing future interaction. Effective prediction therefore requires a representation of what the robot has learned about the evolving environment.

Excavation makes this challenge concrete. A scoop removes material, changes support, and alters where the next scoop should enter. Learned wheel-loader models already predict pile evolution and loading performance \citep{aoshima,aoshima2025}. During repeated excavation, however, the sensing window moves across the pile. A previously disturbed region can disappear from view while its physical consequences continue to affect later actions.

\begin{figure}[t]
\centering
\includegraphics[width=\linewidth]{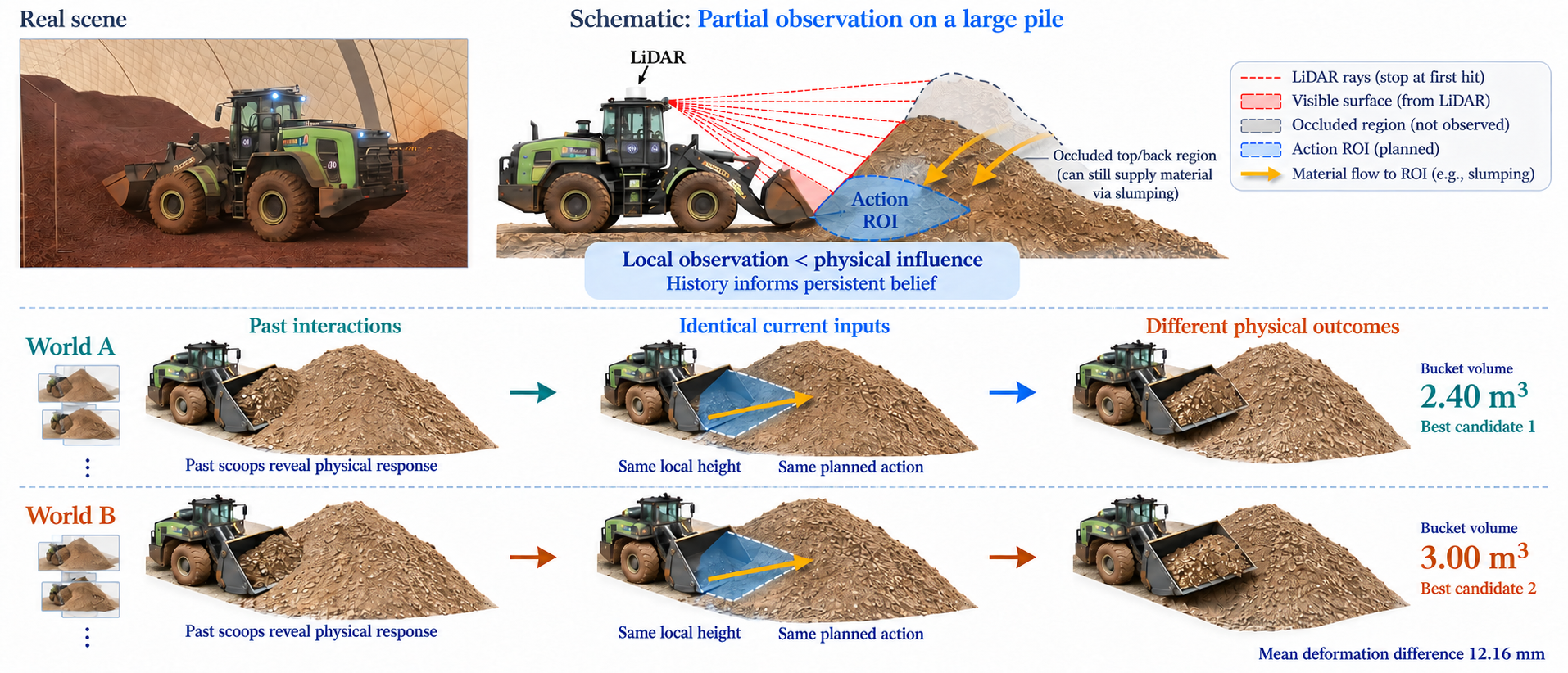}
\caption{Local sensing and physical ambiguity. The upper panel illustrates hidden influence beyond the visible surface. The controlled numerical pair below has identical current inputs but different outcomes and best actions.}
\label{fig:ambiguity}
\end{figure}

Consider the controlled pair in Figure~\ref{fig:ambiguity}. Worlds A and B have numerically identical current local height maps and planned actions. The same scoop nevertheless loads 2.40~$\mathrm{m^3}$ in A and 3.00~$\mathrm{m^3}$ in B. Their next deformation fields differ by 12.16~mm on average, and their best candidates differ. A deterministic predictor with identical current inputs produces identical predictions. Past interactions supply the evidence needed to distinguish these physically different worlds.

Our central idea is to treat completed interactions as measurements of persistent physical state. Before/after geometry reveals where material moved, while measured load and force reveal how the pile responded. These signals have complementary structure. Deformation belongs to world locations, and physical response provides context for interpreting a new action. Retaining both lets a world model revise its predictions as experience accumulates.

\method{} organizes this evidence into spatial and response memories. A candidate action reads the relevant registered deformation and response context, which correct an observation-conditioned prior through bounded residuals. Execution writes measured evidence, while rollout maintains a separate imagined state. The learned parameters remain fixed at deployment, and predictive state evolves with interaction. The contribution is this organization and use of physical evidence, with sparse access serving as its implementation interface.

Our contributions are threefold.
\begin{enumerate}
\item We introduce an interaction-driven world model that resolves local-observation ambiguity through persistent physical belief.
\item We develop a spatial--response state that connects where the environment changed with how it responded to candidate actions.
\item We validate the method through controlled prediction and decision studies, and further demonstrate its generality in Newton/MPM and real excavation.
\end{enumerate}

\section{Related Work}
\subsection{Learning-based excavation and granular prediction}
Learned excavation models operate at both terrain and particle levels. \citet{schenck} predict image-based action effects in granular manipulation. \citet{saku} combine a convolutional autoencoder with an LSTM to predict soil deformation from measured excavation sequences. For wheel loaders, \citet{aoshima} predict pile shape and loading performance, and \citet{aoshima2025} integrate these models with look-ahead search for loading sequences and transportation costs. Graph-based simulators model interacting particles \citep{gns}, including granular--rigid-body dynamics for manipulation \citep{tuomainen}. Under local sensing, visible geometry leaves support and material response partially hidden. \method{} retains the physical evidence that completed interactions reveal about these conditions.

\subsection{World models and interaction-based adaptation}
PlaNet learns latent dynamics for planning \citep{planet}, while Dreamer and DreamerV3 learn behavior through imagined trajectories \citep{dreamer,dreamerv3}. TD-MPC2 combines task-oriented world models with trajectory optimization \citep{tdmpc2}. Generative models additionally synthesize controlled visual futures, including risk-conditioned multi-view driving scenarios \citep{lin2026risk}. Interaction-based adaptation extracts predictive context from past transitions. PEARL infers probabilistic task context \citep{pearl}, and CoDeGa adapts granular scooping predictions to material and terrain changes \citep{codega}. These approaches motivate history-conditioned prediction, while leaving room for explicit correspondence between physical evidence and the world locations of future actions. \method{} provides this correspondence through coupled spatial and response state.

\subsection{Spatial and sequence memory under partial observation}
Neural Map introduces writable spatial memory \citep{neuralmap}, Cognitive Mapping and Planning accumulates spatial belief for navigation \citep{cmp}, and MapNet maintains allocentric memory through localization and registration \citep{mapnet}. Convolutional recurrence preserves spatial features over time \citep{convgru}, while attention provides content-dependent access to stored representations \citep{transformer,smt}. These mechanisms establish useful ways to retain information beyond the current view. \method{} gives memory a physical transition role by recording action consequences and reading them in the frame of a proposed interaction.

\section{Problem Formulation}
Let $s_t$ be the hidden pile state and $o_t$ the robot's local observation before action $a_t$. A completed scoop yields $y_t=(\Delta H_t,\mathbf r_t)$, where $\Delta H_t$ is signed terrain change and $\mathbf r_t=[V_t,F_t,q_t,R_t]$ contains bucket volume, peak force, boundary-flow volume, and failure-moved volume. Flow and failure quantify distinct forms of redistribution in cubic meters. Domain-specific evaluations use their measured channels.

The history $\mathcal H_t=\{(o_i,a_i,y_i)\}_{i<t}$ contains completed events. We learn a predictor with a persistent state $B_t$,
\begin{equation}
\hat y_t=f_\theta(o_t,a_t,B_t),\qquad
B_{t+1}=\mathcal U(B_t,o_t,a_t,y_t).
\label{eq:world}
\end{equation}
The learning objective is to retain distinctions in $\mathcal H_t$ that predict outcomes beyond the current observation. Here, physical belief denotes a history-conditioned predictive state. Measurements update it after execution. Hypothetical predictions update a separate imagined copy. All candidates at a decision step read the same pre-action state, and future measurements enter only after the corresponding action is completed.

\section{PileBelief World Model}
\begin{figure}[t]
\centering
\includegraphics[width=\linewidth]{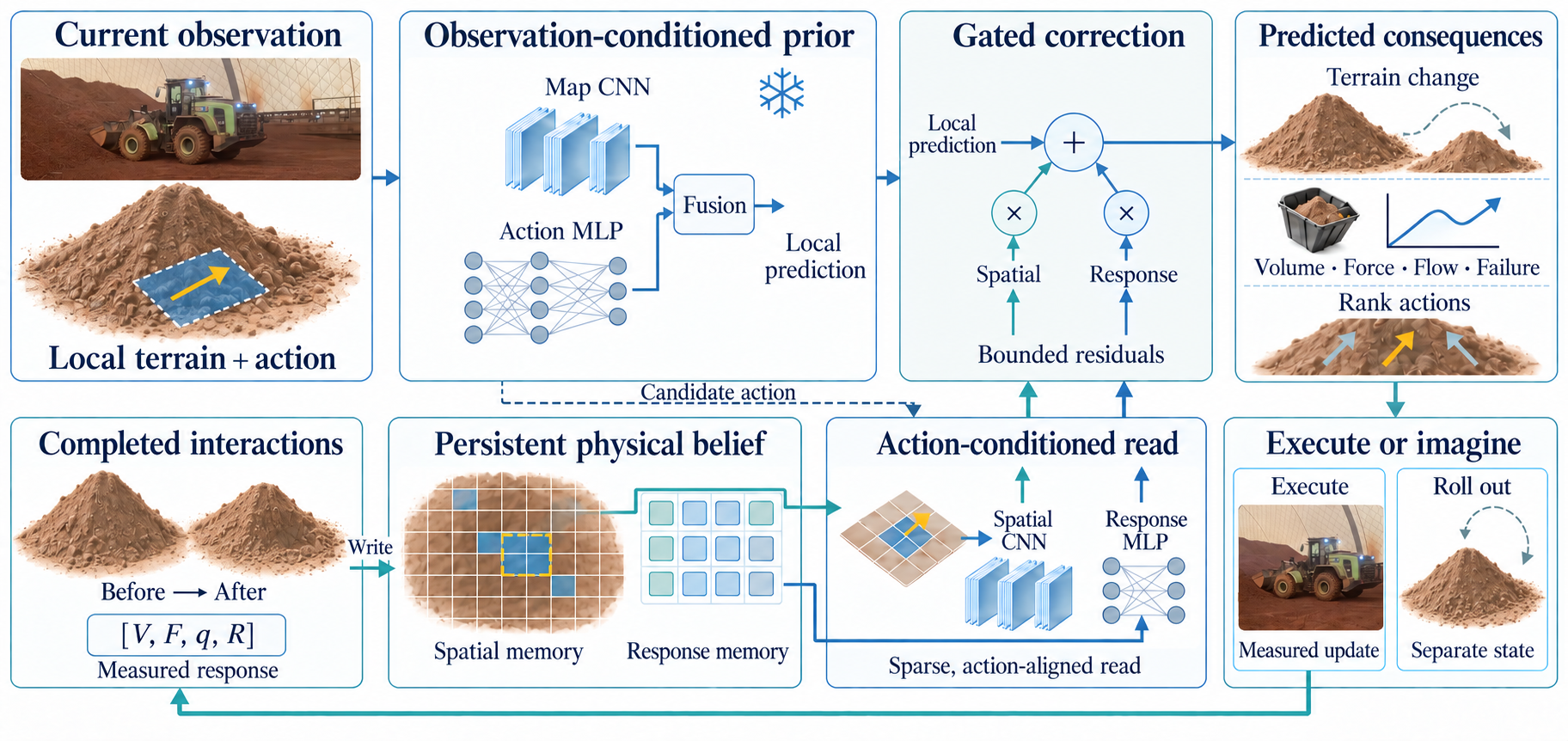}
\caption{PileBelief workflow. Completed interactions update spatial and response memories. Action-aligned reads provide gated corrections to a frozen local prior. Execution updates measured belief, while rollout advances a separate imagined state.}
\label{fig:architecture}
\end{figure}

Figure~\ref{fig:architecture} shows how measured interactions update belief and how proposed actions read it for prediction.

\subsection{Observation-conditioned physical prior}
The local operator encodes an action-centered height crop, a validity mask, and planned sweep geometry with a CNN. An MLP embeds heading, travel, depth, speed, and curl. Fused features produce a terrain-change map and scalar physical outcomes. Entry position determines the crop and world registration. We first train this observation-conditioned prior $\hat y_t^L$ and then freeze it during memory-residual training. It anchors prediction to the current surface. An empty history disables both correction branches and exactly recovers the local operator. Appendix~\ref{app:implementation} specifies dimensions and normalization.

\subsection{World-addressed spatial and response state}
A completed interaction supplies signed post-minus-pre deformation, its action-to-world transform, and measured outcomes. We register deformation in world cells, average same-event collisions, and use validity masks to identify supported writes. Successive deformations advance the world-height state additively. New observations refresh visible cells, while other cells retain their state. This mechanism preserves physical consequences as the local sensing window moves.

The state separates accumulated geometry from the evidence read by the correction network. The reference transition configuration stores the three most recent deformation events in chronological channels of a registered map $G_t$, together with three outcome vectors in $Q_t\in\mathbb R^{3\times4}$. Thus, recent event storage is bounded, while the world-height state accumulates effects over time. Spatial memory retains where changes occurred, and response memory summarizes the measured material--machine behavior. A recursive-state control additionally replaces replay of older events with a masked spatial ConvGRU, allowing explicit tests of long-term retention beyond a raw-event cache (Section~\ref{sec:persistence}).
Appendix~\ref{app:mpm} specifies its multi-material MPM adaptation.

\subsection{Action-aligned access and gated correction}
For candidate action $a$, the reader selects historical tiles and resamples their evidence into the candidate's longitudinal--lateral frame,
\begin{equation}
E_t(a)=\mathcal W_a\!\left(G_t[\operatorname{TopK}(\boldsymbol\rho_t)]\right).
\label{eq:query}
\end{equation}
Here $\rho_t$ scores valid tiles by mean absolute deformation, and $\mathcal W_a$ performs world-to-action resampling. The reference configuration retains at most 12 tiles. The tile selection is evidence-based, and the resampling is action-conditioned. This sparse reader bounds the historical input to the correction network.

A spatial CNN encodes $E_t(a)$, and a response MLP encodes the normalized chronological outcomes. Their features and the current crop features parameterize separate corrections,
\begin{equation}
\hat y_t=\hat y_t^L+g_t^S\odot\delta_t^S+g_t^R\odot\delta_t^R.
\label{eq:residual}
\end{equation}
Sigmoid gates control the contribution of each branch, and bounded residuals limit the magnitude of its correction. The network can therefore adjust a prediction using relevant evidence while preserving a current-observation anchor. Keeping weights fixed and intervening on evidence directly tests whether spatial location, action alignment, and physical response affect its predictions.

\subsection{Learning and imagined state updates}
Training combines normalized valid-pixel height $L_1$ loss, smooth-$L_1$ outcome loss, and a matched-group loss,
\begin{equation}
\mathcal L=\mathcal L_{\Delta H}+\mathcal L_{\mathbf r}+4\mathcal L_{\mathrm{matched}}.
\label{eq:loss}
\end{equation}
Within groups sharing current observation and action, $\mathcal L_{\mathrm{matched}}$ compares mean-centered predictions and targets. It emphasizes the physical differences explained by history. The residual stage alternates dense and sparse access to the same evidence. Rollout training additionally exposes the model to detached predicted inputs.

During imagination, predicted deformation advances the copied height state and predicted outcomes enter its history. Measured belief remains available for subsequent real observations. The evaluated rollout starts with the local prior and enables correction after the first imagined write. Consequently, first-step equality to Local is a property of this protocol, and later steps evaluate correction as predictions reshape the imagined state.

\subsection{Candidate action evaluation}
A separately trained scoring head reads frozen world-model features and candidate context. It combines a local utility score with a gated history-conditioned residual and a calibrated force/failure penalty. Training uses utility regression, best-candidate classification, auxiliary outcome prediction, and paired correct/wrong-history losses. The selected action maximizes the predicted score over a shared feasible candidate set. This supplies an offline decision interface to belief, with completed outcomes serving as training and evaluation labels. Appendix~\ref{app:decision} gives the scoring objective and utility.

\section{Experiments}
\subsection{Evaluation setup}
MiniSlope is our reduced-order 2.5D excavation simulator. It combines geometric removal with volume-conserving critical-slope relaxation, following the cellular-automaton formulation of \citet{spinelli}. Our pipeline adds material-response variation, localized failure, and boundary-flux accounting. Controlled panels match current geometry and action while varying hidden response and connectivity. Newton/MPM and field measurements provide distinct dynamics and sensing settings.

We report height and physical-outcome MAEs. The controlled joint score is
\begin{equation}
J=\frac12\left(\frac{e_H}{0.08~\mathrm m}
+\frac14\sum_{j=1}^{4}\frac{e_{r_j}}{c_j}\right),\quad
\mathbf c=[1.5~\mathrm{m^3},100~\mathrm{kN},0.5~\mathrm{m^3},0.5~\mathrm{m^3}].
\label{eq:joint}
\end{equation}
Five-step error $J_{1:5}$ averages all five prediction steps. Regret is the realized utility gap between the selected and oracle-best candidates. Model-level comparisons use matched information, data partitions, and evaluation settings within each task. Dataset counts, training details, and the seed coverage of each study appear in Appendix~\ref{app:protocols}.

\subsection{Comparison with history-based world models}
\label{sec:comparison}
Table~\ref{tab:models} compares Local, ConvGRU, History Transformer, and \method{} on one-step prediction, five-step rollout, and action selection. The history-based methods receive the same completed deformation, executed actions, spatial transforms, validity masks, and measured outcomes. ConvGRU encodes this stream recurrently, while the Transformer attends to spatial event tokens with time and pose embeddings. All methods use matched task-specific data partitions, supervision, and decision-readout settings. The three history modules share the frozen local prior and use matched training budgets, allowing the comparison to focus on how interaction evidence is organized and used.

\begin{table}[!htb]
\caption{Matched-history model comparison.}
\label{tab:models}
\centering\tablesize\setlength{\tabcolsep}{5pt}
\begin{tabular}{@{}lccc@{}}
\toprule
Model & $J_1\downarrow$ & $J_{1:5}\downarrow$ & Regret$\downarrow$\\
\midrule
Local & 0.5404 & 0.9408 & 0.2757\\
ConvGRU & 0.4698 & 0.8891 & 0.1432\\
History Transformer & 0.4575 & 0.8710 & 0.1346\\
\textbf{PileBelief} & \textbf{0.4202} & \textbf{0.8393} & \textbf{0.0950}\\
\bottomrule
\end{tabular}
\end{table}

Both history baselines improve on current-only prediction. \method{} achieves the lowest error and regret, reducing one-step and five-step joint error by 8.2\% and 3.6\% relative to History Transformer, and regret by 29.4\%. These gains under matched inputs and training budgets show that organizing interactions into world-addressed deformation and physical-response state improves prediction and action evaluation.

\subsection{Information and physical evidence attribution}
\label{sec:mechanism}
Cumulative input ablations show that completed history reduces joint error by 71.8\% beyond all explicit current-view priors (Appendix~\ref{app:priors}). The largest geometric-prior gain comes from the action envelope. Further geometric and mass/flow features give smaller gains. Completed responses thus supply information about physical conditions that current-view features leave unresolved.

\begin{table}[t]
\caption{Fixed-network memory interventions.}
\label{tab:mechanism}
\centering\tablesize\setlength{\tabcolsep}{3.2pt}
\begin{tabular}{@{}lcccccc@{}}
\toprule
Evidence & $H$ (mm)$\downarrow$ & $V$ (m$^3$)$\downarrow$ & $F$ (kN)$\downarrow$ & $q$ (m$^3$)$\downarrow$ & $R$ (m$^3$)$\downarrow$ & $J\downarrow$\\
\midrule
Local / empty & 46.54 & 1.0875 & 55.33 & 0.0735 & 0.2858 & 0.5404\\
Response & 49.44 & 0.9297 & 33.55 & 0.0804 & 0.2332 & 0.5068\\
Sparse spatial & 45.92 & 0.9327 & 28.41 & 0.0689 & 0.2096 & 0.4699\\
Response + dense & \textbf{42.02} & 0.8901 & 23.61 & \textbf{0.0390} & 0.1770 & 0.4203\\
\textbf{PileBelief} & 42.11 & \textbf{0.8887} & \textbf{23.31} & 0.0392 & \textbf{0.1762} & \textbf{0.4202}\\
\midrule
Shifted spatial & 48.81 & 0.9203 & 32.81 & 0.0673 & 0.2263 & 0.4962\\
Misaligned action & 48.14 & 0.8897 & 24.16 & 0.0395 & 0.1771 & 0.4594 \\
Wrong response & 42.33 & 1.3478 & 55.66 & 0.0630 & 0.3109 & 0.5400\\
\bottomrule
\end{tabular}
\end{table}

Table~\ref{tab:mechanism} fixes network weights and intervenes on evidence. Combining spatial and response memory outperforms either branch alone and reduces joint error by 22.2\% relative to Local. Spatial memory captures where the pile changed, while response memory captures how it responded. Crucially, \method{} matches Response + dense spatial with 12 selected historical tiles, compared with 36.2 on average for dense access, preserving predictive accuracy with substantially sparser historical reads. Spatial shifts, action misalignment, and incorrect response history all increase joint error. Their different effects across outputs reveal complementary roles for where an interaction occurred and how the material responded.

\subsection{Persistent memory and multi-step prediction}
\label{sec:persistence}
\begin{figure}[t]
\centering
\includegraphics[width=\linewidth]{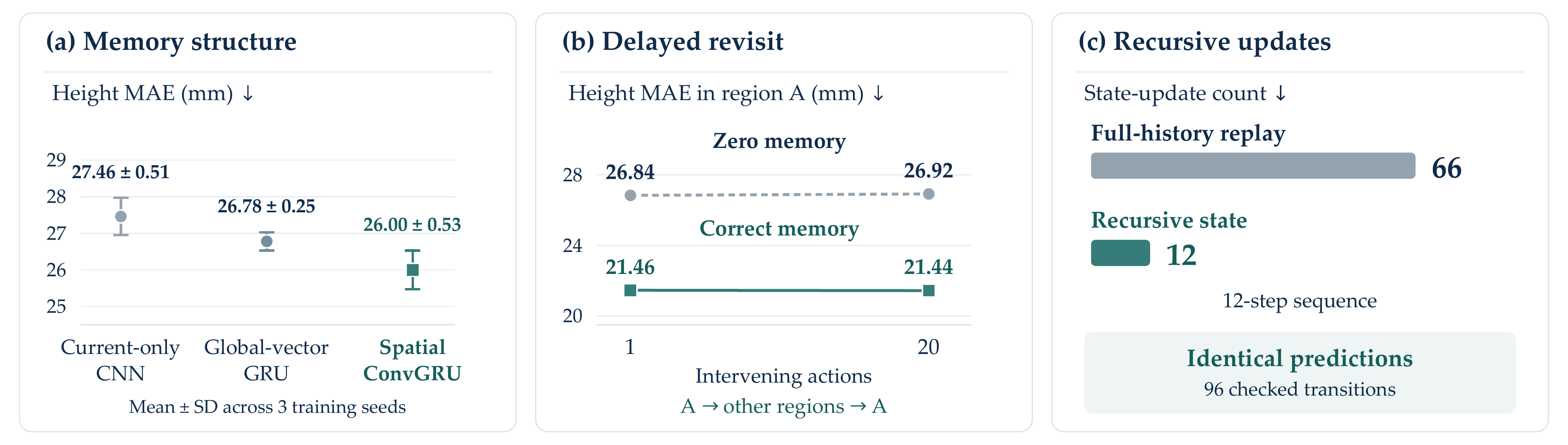}
\caption{Persistent memory. (a) Three-seed structure controls. (b) Measured delayed-revisit endpoints. (c) Recursive and full-replay update counts.}
\label{fig:persistence}
\end{figure}

Three-seed memory-structure controls compare current-only features, a global recurrent summary, and spatial ConvGRU state under a common super-network and training protocol (Figure~\ref{fig:persistence}a). Spatial recurrence achieves the lowest mean height error. Correct memory also retains its benefit after 20 intervening actions in the recorded revisit study (Figure~\ref{fig:persistence}b). These results support keeping physical evidence associated with the affected world region as the robot interacts elsewhere.

The recursive control stores one recent raw event and compresses older evidence into spatial state. Its predictions match full replay on 96 checked transitions, with 12 recursive updates instead of 66 replay updates over a 12-step sequence. Removing the older state increases endpoint height error by $8.12\pm1.68$~mm across three seeds. Persistence therefore preserves useful evidence beyond a latest-event cache.

\begin{figure}[t]
\centering
\includegraphics[width=\linewidth]{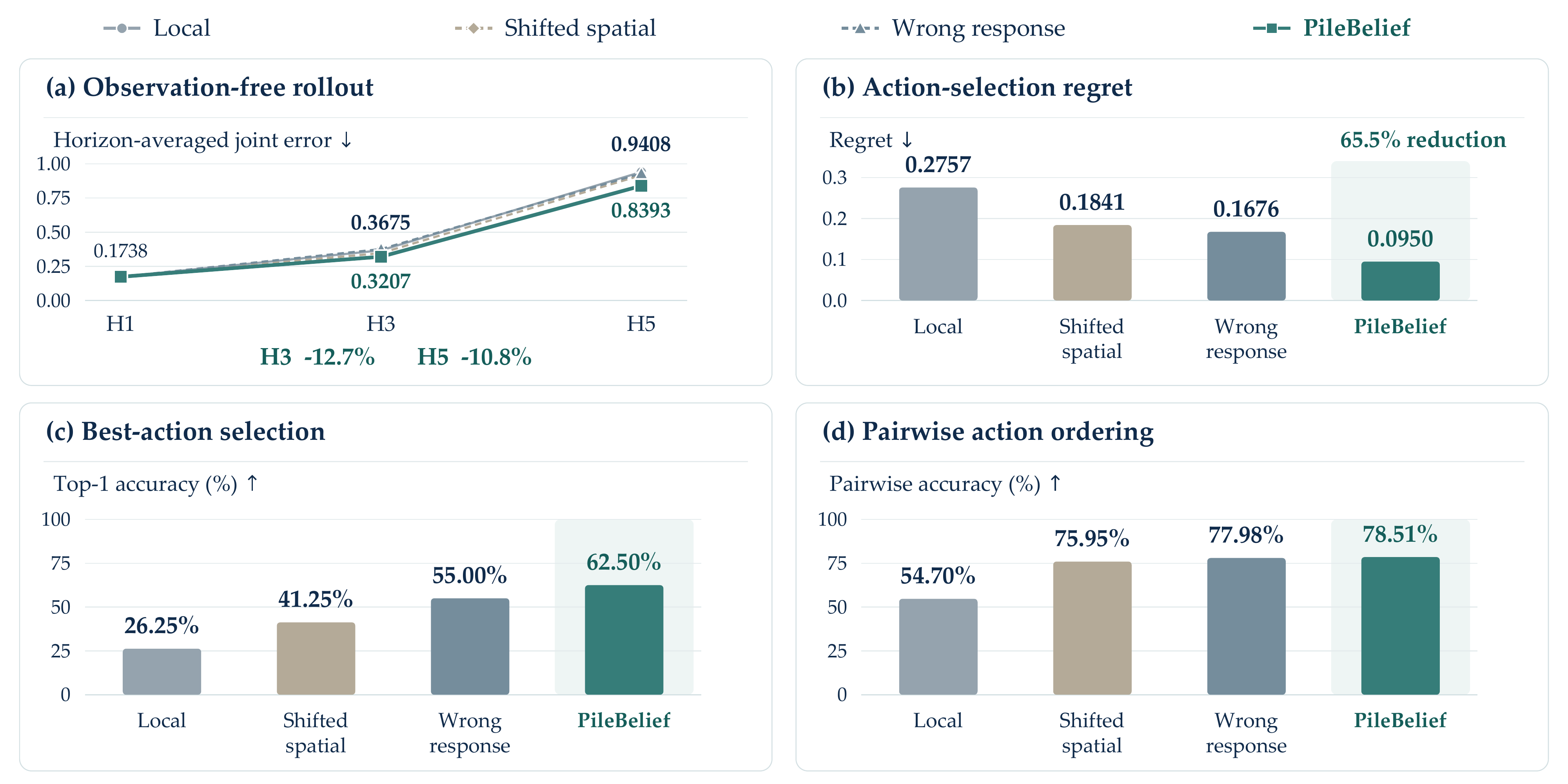}
\caption{Prediction and decisions in the recorded controlled studies. (a) Free rollout. (b--d) Regret, Top-1 accuracy, and pairwise ordering with seven candidates per state.}
\label{fig:prediction_decision}
\end{figure}

For observation-free rollout, 64 fresh five-step episodes start from measured observations and subsequently consume predicted heights, outcomes, and masks. Figure~\ref{fig:prediction_decision}a reports error averaged through each horizon. Belief reduces joint error by 12.7\% through three steps and 10.8\% through five steps. An independent three-seed height-only control reduces five-step MAE from $93.56\pm3.46$ to $89.52\pm1.69$~mm, with a gain in every seed. Physical state remains useful when the model advances through its own predicted consequences.

The recorded decision study uses 240 training, 80 calibration, and 80 validation states, with seven feasible candidates per state. All candidates share the same pre-action pile. PileBelief reduces validation regret by 65.5\% relative to Local and selects the exact best action in 62.50\% of states (Figure~\ref{fig:prediction_decision}b--d). Corrupted spatial or response evidence weakens selection. Interaction history consequently affects action preferences as well as physical prediction. Table~\ref{tab:models} extends the regret comparison to recurrent and attention-based history models under matched decision settings.

\subsection{Memory retention across delayed revisits}
\label{sec:ood}
This study tests whether spatial recurrent state remains useful when the robot revisits an affected region after acting elsewhere. Revisit gaps are 1 and 20 actions, with 20 matching the training maximum. Correct and zeroed memory are evaluated at the same targets with fixed weights.

\begin{table}[!htb]
\caption{Revisit height MAE (mm) across three seeds.}
\label{tab:ood}
\centering\tablesize\setlength{\tabcolsep}{5pt}
\begin{tabular}{@{}lcc@{}}
\toprule
Evidence & Gap 1$\downarrow$ & Gap 20$\downarrow$\\
\midrule
Zero memory & $26.84\pm1.45$ & $26.92\pm1.47$\\
\textbf{PileBelief (correct memory)} & $21.46\pm0.50$ & $21.44\pm0.53$\\
\bottomrule
\end{tabular}
\end{table}

Correct memory reduces height error by 20.4\% at gap 20, with a gain in every seed (Table~\ref{tab:ood}). Its error remains nearly unchanged from gap 1 to 20, showing that registered physical evidence remains useful after intervening actions. Appendix~\ref{app:matched} details the targets, aggregation, and split.

\subsection{Newton/MPM and real excavation}
\label{sec:generality}
Finally, we evaluate the modeling approach under different dynamics and sensing conditions. Both studies train within the target domain and evaluate disjoint sequences. They test the generality of interaction-conditioned prediction across implementations and measurement interfaces.

\subsubsection{History updates under Newton/MPM dynamics}
Using Newton material-point dynamics \citep{newton}, we evaluate an expanded collection of 309 valid transitions under seven-fold material-group holdout. Empty-memory, fixed first-three, and rolling recent-three models are trained from scratch with matched initialization and schedules, using one seed per fold. With fixed evaluation weights, all models predict the same 164 targets from 29 segments and six eligible groups, starting at the fifth scoop. Errors are averaged within 27 scene--seed groups and then equally across materials.

\begin{table}[t]
\caption{Newton/MPM prediction on held-out material groups.}
\label{tab:mpm}
\centering\tablesize\setlength{\tabcolsep}{6pt}
\begin{tabular}{@{}lccc@{}}
\toprule
Condition & Height (mm)$\downarrow$ & Bucket (L)$\downarrow$ & Force (kN)$\downarrow$\\
\midrule
Local (empty memory) & 18.721 & 35.808 & 3.804\\
Fixed first three & 15.228 & 47.602 & 4.064\\
\textbf{PileBelief (recent three)} & \textbf{12.673} & \textbf{33.666} & \textbf{3.653}\\
\bottomrule
\end{tabular}
\end{table}

Recent history reduces height error by 32.3\% relative to Local, improving all six groups (Table~\ref{tab:mpm}). Bucket and force errors decrease by 6.0\% and 4.0\%, respectively. With the same three-event budget, refreshing history reduces bucket error by 29.3\% relative to fixed early evidence, improving every group. This consistency highlights the value of updating physical state as excavation progresses. On the newly collected 106-target subset covering two groups, height, bucket, and force errors fall by 33.1\%, 17.1\%, and 20.0\% relative to Local. These are observation-conditioned one-step predictions. Appendix~\ref{app:mpm} reports protocols, material-level uncertainty, and spatial-coverage diagnostics.

\subsubsection{Real excavation measurements}
\begin{figure}[t]
\centering
\includegraphics[width=\linewidth]{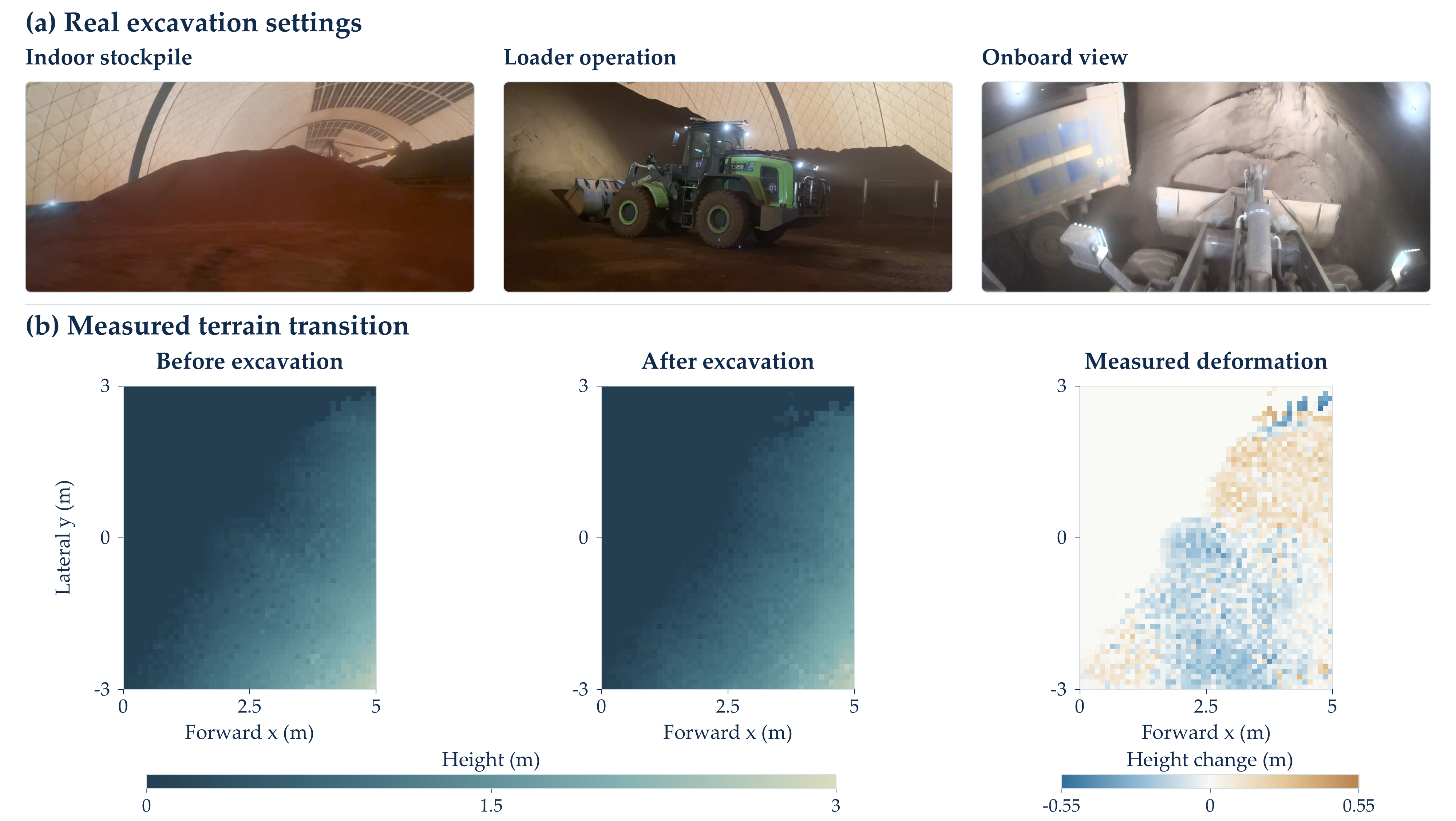}
\caption{Indoor excavation settings (a) and a separately recorded terrain transition (b). Pre/post maps share a height scale. Blue denotes removal and orange deposition.}
\label{fig:real}
\end{figure}

Real sequences contain pre/post pile scans and independent point-cloud estimates of bucket volume. The split contains 71 training targets from 32 continuous segments, 16 development targets from eight segments, and 43 held-out targets from 20 segments. One training seed is used, with model selection confined to development data. Figure~\ref{fig:real} shows physical settings and a separately recorded terrain transition.

The real-domain predictor receives the current pre-excavation crop and up to three completed interactions, encoded in each scoop's local frame. It predicts outcomes under the recorded collection policy using an observation--history interface. Errors are averaged equally across segments. The real-data score equally weights height and bucket errors normalized by $0.1443$~m and $0.2127$~$\mathrm{m^3}$, respectively.

\begin{table}[t]
\caption{Real excavation prediction on held-out segments.}
\label{tab:real}
\centering\tablesize\setlength{\tabcolsep}{5pt}
\begin{tabular}{@{}lccc@{}}
\toprule
Condition & Height (mm)$\downarrow$ & Bucket (m$^3$)$\downarrow$ & $J_{\mathrm{real}}\downarrow$\\
\midrule
Local & 114.74 & 0.2207 & 0.9164\\
Capacity-matched Local & 112.65 & 0.2306 & 0.9324\\
Zero history input & 112.83 & 0.2357 & 0.9451\\
Matched shuffled history & 113.10 & 0.2120 & 0.8903\\
\textbf{PileBelief} & \textbf{112.17} & \textbf{0.2056} & \textbf{0.8720}\\
\bottomrule
\end{tabular}
\end{table}

\method{} achieves the lowest mean error across all three metrics (Table~\ref{tab:real}). Relative to Local, bucket and joint errors decrease by 6.8\% and 4.8\%. Correct history also improves on capacity-matched, zero-history, and matched-shuffle controls. These results extend the predictive value of completed interactions to measured excavation outcomes, particularly load estimation. Appendix~\ref{app:real} specifies the controls and data interface.

\section{Discussion}
\label{sec:discussion}

Our results support a state-design principle for world models of robot-modified environments. Interaction history becomes more useful when its physical meaning is preserved. Comparisons with recurrent and attention-based models show the value of organizing the same evidence into spatial deformation and physical response. Evidence interventions further connect this benefit to correspondence among world location, candidate action, and measured outcome. The comparable performance of sparse and dense access suggests that informative state can be exposed through a compact query interface.

This perspective separates learning a transition operator from maintaining knowledge of the evolving environment. Fixed model weights provide reusable predictive structure, while completed interactions update the state on which predictions depend. The equal-budget MPM comparison shows the value of refreshing this evidence during excavation. In the larger MPM scenes, masking history outside the current view retains most of the height-prediction gain, suggesting that memory also tracks evolving local physical conditions (Appendix~\ref{app:mpm_coverage}). Separate measured and imagined states preserve executed evidence during candidate evaluation.

The current design relies on spatial registration and the coverage of past interactions. Robustness to registration drift and uncertainty-aware evidence retention are therefore important next steps. Newton/MPM and real excavation establish the value of history under target-domain training. Transferring a shared model across domains and evaluating its action choices in closed-loop excavation would further extend the practical reach of persistent physical belief.

\FloatBarrier
\section{Conclusion}
\method{} addresses current-view ambiguity in robotic excavation by retaining physical evidence revealed by completed interactions. World-addressed deformation and response memory refine an observation-conditioned prior through action-aligned, gated corrections. Experiments show that this state resolves hidden physical differences, remains useful across delayed revisits, and improves observation-free prediction and offline action selection. Newton/MPM and real excavation further demonstrate the generality of interaction-conditioned prediction across dynamics and sensing settings. With fixed deployment weights and an evolving physical belief, \method{} provides a foundation for world models of environments that robots continually reshape.
\label{sec:conclusion}

\subsection*{AI use statement}
\label{sec:ai_use}
Codex assisted experimental design, literature organization, language revision, and reference formatting. The authors take responsibility for the final scientific claims, text, and artifacts.

\subsection*{Ethics statement}
\label{sec:ethics}
Excavation involves heavy machinery and potentially unstable material. The reported action-selection results are offline evaluations. Deployment should combine predictive models with independently validated motion constraints, force limits, site-specific risk assessment, and operator oversight.

\subsection*{Reproducibility statement}
\label{sec:reproducibility}
Section~4 and Appendix~\ref{app:implementation} specify the model, state updates, losses, and training environment. Appendix~\ref{app:protocols} describes data partitions, metrics, and statistical reporting. Appendix~\ref{app:matched} specifies the matched baselines and supporting mechanism studies. Appendix~\ref{app:decision} gives the offline utility and ranking interface. Appendices~\ref{app:mpm} and \ref{app:real} provide the Newton/MPM and real-data protocols, controls, and supporting analyses. Code and evaluation scripts will be publicly released upon acceptance.

\bibliography{references}
\bibliographystyle{iclr2027_conference}

\clearpage
\appendix
\section{Model implementation and training}
\label{app:implementation}
\subsection{Local observation and action encoding}
The reference transition operator uses a $5\,\mathrm m\times5\,\mathrm m$ action-centered crop on a $33\times25$ grid. Its five channels are anchor-relative height divided by $3$~m, validity, sweep weight, depth-weighted sweep, and along-track progress. The depth raster scales nominal depth by $1.5$~m, and progress is clipped to $[0,1]$ within the sweep. A three-layer $3\times3$ CNN has widths $5\!\rightarrow\!24\!\rightarrow\!24\!\rightarrow\!24$ and SiLU activations.

The six action features are heading sine and cosine, travel divided by $5$~m, depth divided by $1.2$~m, speed divided by $2.5$~m/s, and curl divided by $60$ degrees. A $6\!\rightarrow\!24\!\rightarrow\!24$ MLP broadcasts its embedding over the crop. Entry position determines crop extraction and registration. A convolutional head predicts signed terrain change, and a pooled-feature head predicts scalar outcomes. Predicting bucket volume separately from surface loss accommodates material flow across the crop boundary.

\subsection{Persistent state and evidence reads}
The reference world map spans $30\,\mathrm m\times30\,\mathrm m$, with $0.25$-m cells and $1$-m tiles. The rigid action-to-world transform assigns each valid deformation sample to its nearest world cell. Samples of the same event assigned to one cell are averaged. Successive event deformations advance accumulated world height, and incoming observations refresh supported visible geometry.

The accumulated height map, recent deformation channels, and response FIFO have different roles. The map retains evolving geometry. The three chronological deformation channels retain recent interaction evidence with world addresses. The response FIFO contains the latest three completed $[V,F,q,R]$ vectors. These recent-event buffers have finite length. The recursive-state controls instead compress older registered event features into a masked spatial ConvGRU state, with updates restricted to supported cells. This distinction specifies the state used by each evaluation without treating all memory storage as an identical mechanism.

For reference transition reads, an active tile contains at least one supported cell and is scored by mean absolute deformation across its cells and event channels. At most $K=12$ tiles are retained, unselected locations are zeroed, and world-to-action sampling produces $E_t(a)\in\mathbb R^{3\times33\times25}$. The implementation refers to this reader/write interface as Sparse Query and Evidence Writing.

\subsection{Correction networks and training schedule}
The aligned three-channel deformation input is divided by $0.08$~m and encoded with a $3\!\rightarrow\!24\!\rightarrow\!24$ CNN. Response vectors use the four physical scales in Equation~\eqref{eq:joint}, are flattened chronologically, and pass through a $12\!\rightarrow\!24\!\rightarrow\!24$ MLP. Gates are $0.35\sigma(Wz+b)$ with zero-initialized weights and bias $-2.5$. Spatial and response terrain residuals are bounded by $0.07\tanh(\cdot)$ and $0.05\tanh(\cdot)$ meters, respectively. Scalar residuals act in normalized units with $\tanh$ bounds.

The local stage uses AdamW at $8\times10^{-4}$ for 35 epochs with batch size 64. The residual stage uses $6\times10^{-4}$ for 50 epochs with alternating dense and sparse reads, batching eight training groups per update. Weight decay is $10^{-4}$ and gradients are clipped to norm 5. The information ablation trains each arm for 35 epochs with matched maximum input capacity, disabled channels zeroed, and fixed final-epoch selection. Rollout training adds 18 epochs of detached predicted-input exposure at $1.5\times10^{-4}$. Server-side training used NVIDIA A100-SXM4 GPUs with 80~GB memory, PyTorch 2.5.1, and CUDA 12.1.

The matched-group loss centers predictions and targets separately within groups having identical current observation and action. Normalized height $L_1$ and outcome smooth-$L_1$ losses then compare these centered values. The loss trains history-dependent variation around each shared current-input condition. Group construction is confined to each data split.

\subsection{Measured and imagined state management}
At execution time, all candidates read the same pre-action belief. Only the executed action contributes a measured write. During rollout, the model copies the starting state, advances it with predicted terrain change, and appends predicted outcomes to the imagined history. The measured state remains separate. The evaluated provenance switch uses the local prediction until the first imagined write, then enables belief correction. The resulting first-step equality to Local follows directly from the switch.

\section{Data and evaluation protocols}
\label{app:protocols}
The information-ablation panel contains 1,280 training and 320 validation transitions. The registered world-tile panel contains 640 training and 320 validation transitions, including eight validation geometry--position clusters. Geometry and position groups are disjoint across the corresponding training and validation splits. Both panels report one training seed. Their scores measure controlled information and evidence interventions.

The recurrent memory-structure and independent height-rollout studies report three seeds. The joint rollout panel contains 64 fresh five-step episodes. The recorded offline ranking panel contains 240 training, 80 calibration, and 80 validation states. The Newton/MPM collection contains 309 valid transitions from 37 continuous segments, grouped into 35 distinct source/scenario/seed groups. Its primary evaluation uses one training seed per material-group holdout fold and covers 164 targets from 29 segments, 27 scene--seed groups, and six eligible materials. Real excavation reports one seed on 43 held-out targets from 20 continuous segments. Each study uses its stated task, aggregation, and split. Detailed mechanism controls appear in Appendix~\ref{app:matched}, with domain-specific protocols in Appendices~\ref{app:mpm} and \ref{app:real}.

\subsection{Statistical reporting}
Table~\ref{tab:models} reports measured model-level point estimates, with the same task-specific aggregation for all compared methods. The one-step and joint-rollout comparisons retain their recorded single-run estimates, while regret follows the shared decision-readout protocol. For studies with available per-seed records, we compute each seed's complete evaluation metric and report the mean and sample standard deviation across three trained seeds. Model selection, when used, is confined to development data; normalization statistics and frozen initialization are derived from the training partition. The revisit statistics in Table~\ref{tab:ood} are computed from the recorded seed-level results.

\section{Baselines and supplementary mechanism studies}
\label{app:matched}

\subsection{Shared information and training budget}
Every history-based method processes the same sequence of completed events. Each event contains the local observation, deformation, action vector, spatial transform, physical responses, and validity mask. The current query contains the same pre-action crop and candidate action for every method. Material labels and hidden simulator state are excluded from model inputs. Histories remain within their assigned episode or segment, and test target outcomes are available only after their corresponding prediction.

All methods share the current-observation interface, output targets, normalization, and task-specific data partitions. The history-based models use the same frozen local prior and receive matched supervision, including matched-input groups and predicted-input exposure. Training and readout settings are held consistent across methods. The comparison changes how completed events are retained and accessed for prediction.

Local uses only the current input. ConvGRU processes event feature maps sequentially with action, response, time, and spatial-coordinate channels. History Transformer encodes event feature-map tokens with action, response, relative-time, and pose embeddings, and lets a current-query token read the causal event sequence through attention. Spatial event tokens preserve regional detail. Both baselines receive the position and response information supplied to \method{}.

The entire evaluated event stream remains available causally to each history model. Recurrent models carry state across events, and the Transformer retains the corresponding event tokens. Sequence truncation and resets use a common protocol.

\subsection{Three primary outputs}
The single-step score is Equation~\eqref{eq:joint} on held-out transitions. Five-step prediction starts from the same measured information for every method and then uses each model's own predicted inputs. The primary rollout metric averages joint error over steps $1{:}5$, with per-horizon curves and per-output errors retained as diagnostics. Teacher forcing is confined to the designated training protocol.

Regret uses the same seven candidates per state and the utility in Equation~\eqref{eq:utility}. Each model receives a separately trained scoring head with the same architecture, labels, optimization budget, and calibration procedure. Candidate scores are computed from pre-action features. The world-model and readout pipeline is finalized before the evaluation candidate states are scored. Top-1 accuracy and pairwise ordering are secondary outputs.

The scored feature dimension is standardized through a learned projection. The frozen-feature and ensemble settings are matched across methods. The decision projection and scoring head belong to the separately trained readout. Figure~\ref{fig:prediction_decision} supplies the spatial- and response-evidence interventions for this decision interface.

\subsection{Cumulative information ablation}
\label{app:priors}
Every arm receives the planned action vector and has the same maximum input capacity and training budget. Disabled input channels are zeroed. The cumulative comparison isolates which deployment-available information contributes to physical prediction.

\begin{table}[h]
\caption{Cumulative deployment-information ablation.}
\label{tab:priors}
\centering\tablesize\setlength{\tabcolsep}{5pt}
\begin{tabular}{@{}lcccc@{}}
\toprule
Inputs & $J\downarrow$ & Height (mm)$\downarrow$ & Bucket (m$^3$)$\downarrow$ & Force (kN)$\downarrow$\\
\midrule
Height & 0.7074 & 70.08 & 1.1049 & 55.41\\
+ gradient & 0.7064 & 69.93 & 1.1046 & 55.41\\
+ action envelope & 0.5386 & 43.46 & 1.0985 & 55.41\\
+ sweep volume & 0.5356 & 43.06 & 1.0966 & 55.41\\
+ mass/flow prior & 0.5287 & 41.78 & 1.0994 & 55.41\\
\textbf{+ completed history} & \textbf{0.1491} & \textbf{10.50} & \textbf{0.2160} & \textbf{15.34}\\
\bottomrule
\end{tabular}
\end{table}

The first five conditions have a force MAE of 55.41~kN. The cumulative geometric inputs substantially improve terrain prediction, while the force output changes with completed physical-response evidence. The joint score combines normalized outputs, so small non-monotonic changes in an individual output can coexist with a lower overall score. This panel attributes the major information gain to completed interaction, with a distinct split and protocol from the fixed-network evidence interventions.

\subsection{Persistent-state controls}
The three-seed current-only, global recurrent, and spatial ConvGRU controls obtain height errors of $27.46\pm0.51$, $26.78\pm0.25$, and $26.00\pm0.53$~mm. These are memory-structure controls under a common super-network. They examine spatial organization at the height-prediction level and complement the complete model-level comparison in Table~\ref{tab:models}.

At delayed revisit endpoints of one and 20 intervening actions, zero-memory height errors are $26.84\pm1.45$ and $26.92\pm1.47$~mm, compared with $21.46\pm0.50$ and $21.44\pm0.53$~mm using correct memory. These are means and sample standard deviations across three trained seeds, each evaluated on four paired target cases per gap in the recorded diagnostic validation split. The reported target region matches Figure~\ref{fig:persistence}b. The 20-step revisit gap matches the maximum gap used during training. The connecting line links measured endpoint means. Zero memory is a same-network state intervention, distinct from a separately trained Local model.

The recursive-state control keeps one recent raw patch and integrates older information into a spatial recurrent state. It reproduces replay predictions on 96 tested transitions. Keeping only the latest event raises the ten-interaction endpoint height MAE by $8.12\pm1.68$~mm. Counting all recurrent updates over 12 sequential predictions gives 12 for incremental state and 66 for repeated prefix replay.

\subsection{Observation-free rollout scores}
The joint rollout averages errors over all steps through the indicated horizon. Local obtains $J_{1:1}=0.1738$, $J_{1:3}=0.3675$, and $J_{1:5}=0.9408$. \method{} obtains 0.1738, 0.3207, and 0.8393. Spatially shifted evidence gives 0.3471 and 0.9173 through three and five steps, while incorrect response history gives 0.3765 and 0.9305. Subsequent heights, responses, and validity masks come from the evolving predicted state.

The independent three-seed height-only rollout control uses its own initialization and feedback protocol. It supplies the reported five-step height MAEs of $93.56\pm3.46$~mm for Local and $89.52\pm1.69$~mm for belief. Its bucket route is shared between the compared conditions. This control quantifies terrain prediction, while the joint rollout panel evaluates the normalized multi-output score.

\section{Candidate action evaluation}
\label{app:decision}
The recorded decision score combines a current-input score with a bounded history-conditioned residual and a calibrated force/failure penalty,
\begin{equation}
S_\psi(o_t,a,B_t)=S^L(o_t,a)+g^{\mathrm{dec}}\bigl(d^{\mathrm{dec}}-p^{\mathrm{risk}}\bigr).
\end{equation}
The history correction uses signed elementwise interactions between projected history and action-context features. Auxiliary outcome heads predict load, force, and failure and support calibration. Training combines utility regression, best-candidate classification, auxiliary outcome prediction, and paired losses contrasting correct and incorrect response evidence. Frozen world-model features support this decision head. The existing readout uses its recorded frozen-backbone feature ensemble, and its scores are reported only for that decision protocol.

The utility for simulator outcome $y$ and proposed action $a$ is
\begin{equation}
U(y,a)=\frac{V}{1~\mathrm{m^3}}-0.22\frac{F}{100~\mathrm{kN}}
-0.18\frac{R}{1~\mathrm{m^3}}-0.055\frac{\ell}{1~\mathrm m}-0.04\frac{v}{1~\mathrm{m/s}},
\label{eq:utility}
\end{equation}
where $\ell$ is planned travel and $v$ is planned speed. This dimensionless objective rewards payload and penalizes force demand, failure, travel, and speed. The seven feasible candidates come from a frozen generator with fixed parameter offsets. Simulator outcomes supply utility labels and offline evaluation, while selection uses pre-execution features.

The recorded validation results are regret 0.2757, Top-1 accuracy 26.25\%, and pairwise accuracy 54.70\% for Local. Shifted spatial evidence yields 0.1841, 41.25\%, and 75.95\%. Incorrect response history yields 0.1676, 55.00\%, and 77.98\%. Correct belief yields 0.0950, 62.50\%, and 78.51\%. These values describe the validation study illustrated in Figure~\ref{fig:prediction_decision}. Under the matched decision settings, ConvGRU and History Transformer obtain regrets of 0.1432 and 0.1346, respectively, completing the model-level comparison in Table~\ref{tab:models}.

\section{Newton/MPM protocols and detailed results}
\label{app:mpm}
\subsection{Multi-material data and evaluation}
The continuous multi-material collection contains 338 scoop records from 37 segments, with 309 valid regression transitions. The remaining 29 records comprise 28 incomplete-retreat failures and one manual interruption. They are retained outside regression evaluation, with missing outcomes left unfilled. Segments sharing the same source, scenario, and scene seed are grouped together, yielding 35 groups. Histories contain valid, completed interactions from the same segment and end before the target action.

Seven jointly varied parameter groups are identified by reference densities of 1400, 1475, 1550, 1625, 1700, 1775, and 1850~$\mathrm{kg/m^3}$. Density, friction, stiffness, and Poisson ratio co-vary, so the density labels identify complete parameter groups. Their valid-transition counts are 191, 43, 23, 16, 14, 11, and 11. Each fold holds out every segment of one group and trains on the other six. Normalization uses training-group statistics, and evaluation uses the final training epoch without held-out model selection. The 164 primary targets begin at scoop five and cover 29 segments, 27 scene--seed groups, and six eligible materials. The 1775 group has no valid target at this stage. Errors are averaged within each scene--seed group and then equally over scene groups within each material and over materials.

Bucket labels are reference volumes obtained by dividing the mass of active particles in the counting region by reference density. The proxy counting-box volume is $0.2656~\mathrm{m^3}$, a geometric quantity distinct from calibrated bucket-cavity capacity. The collection controller uses a density-dependent target load. Masking this parameter in model inputs removes the direct cue, while collection-policy dependence remains part of the dataset. Added records concentrate on the 1400 and 1475 groups, and some sources retain only longer segments. Accordingly, the collection supports prediction comparisons on its defined targets rather than an estimate of the overall excavation failure rate.

\subsection{History windows and target-domain implementation}
For scoop indices starting at one, the two completed-history sets are
\begin{equation}
\mathcal H_t^{\mathrm{first}}=\{1,\ldots,\min(3,t-1)\},\qquad
\mathcal H_t^{\mathrm{recent}}=\{\max(1,t-3),\ldots,t-1\}.
\end{equation}
For each target, the recursive state is reset and the selected completed events are replayed chronologically through the same spatial ConvGRU update and query operators. Local uses empty memory. The two history sets coincide at the fourth scoop and begin to differ at the fifth, motivating the primary evaluation panel. This controlled window replay isolates the value of updated evidence, complementing the incremental carry-over study in Section~\ref{sec:persistence}.

The target-domain implementation reuses the recursive prediction, write, update, and spatial-query operators. Its action input contains eight planned-action features and 24 controller parameters. The density-derived target-payload parameter is masked in both current and historical actions. Only commands available before execution enter the current input. The $5$-m observation remains on a $33\times25$ grid. A global coordinate translation places it in the $30$-m world domain with a $32\times32$ grid and a 32-channel ConvGRU state. Valid observation support replaces the unavailable sweep envelope, with the other two envelope channels zeroed. Completed load and force evidence is written only within that observed support.

Local, fixed-history, and recent-history models are trained separately from the same random initialization in each fold, using no pre-trained MiniSlope or MPM weights. Both history arms have the same active architecture and three-event budget. Local retains the prediction architecture with inactive evidence-encoding and recurrent branches. All arms use the same samples and batch order, 80 epochs, batch size 16, AdamW with learning rate $2\times10^{-4}$ and weight decay $10^{-5}$, gradient clipping at norm 5, and final-epoch evaluation. This target-domain schedule is separate from the reference frozen-prior training in Appendix~\ref{app:implementation}. Sequence-balanced MAE terms use scales of $0.1$~m for height, $0.2~\mathrm{m^3}$ for volume, and $20$~kN for peak force. Historical force retains its logarithmic encoding. The base and residual height outputs retain bounds of $0.18$ and $0.12$~m. Evaluation weights remain fixed, and only completed measurements enter history.

The recorded checks reproduce predictions after reloading all 21 trained checkpoints, verify unchanged source-file hashes, and confirm that future or out-of-window outcomes do not affect predictions. At scoop four, each fixed checkpoint produces identical predictions from the two identical support sets. Collection includes unloading, cleaning, and machine reset between scoops, so each target uses its measured pre-action observation. The reported test is single-step prediction conditioned on real completed interactions. Observation-free rollout is evaluated separately in Section~\ref{sec:persistence}.

\subsection{Per-material results and uncertainty}
Table~\ref{tab:mpm_materials} resolves the primary result into its six eligible material groups. Relative to independently trained Local, recent history improves height in all six groups, bucket volume in four, and peak force in four. Relative to fixed early history, recent history improves bucket prediction in every group. This separates the consistent benefit of refreshing evidence from output-dependent gains over current-only prediction.

\begin{table}[!htb]
\caption{MPM errors by held-out material group.}
\label{tab:mpm_materials}
\centering\tablesize\setlength{\tabcolsep}{5pt}
\begin{tabular}{@{}lcccc@{}}
\toprule
Group & History & Height (mm)$\downarrow$ & Bucket (L)$\downarrow$ & Force (kN)$\downarrow$\\
\midrule
1400 & Local & 51.023 & 89.841 & 4.844\\
 & Fixed & 39.841 & 94.580 & \textbf{4.310}\\
 & PileBelief & \textbf{32.312} & \textbf{72.153} & 4.460\\
\midrule
1475 & Local & 16.198 & 26.883 & 4.528\\
 & Fixed & 13.141 & 43.864 & 4.136\\
 & PileBelief & \textbf{12.350} & \textbf{25.094} & \textbf{2.826}\\
\midrule
1550 & Local & 13.778 & \textbf{19.128} & 4.250\\
 & Fixed & \textbf{7.929} & 36.216 & \textbf{2.534}\\
 & PileBelief & 8.869 & 36.194 & 4.249\\
\midrule
1625 & Local & 8.921 & \textbf{15.665} & 4.308\\
 & Fixed & 9.363 & 35.755 & 5.202\\
 & PileBelief & \textbf{7.688} & 21.707 & \textbf{4.089}\\
\midrule
1700 & Local & 7.568 & 28.303 & \textbf{1.707}\\
 & Fixed & 9.031 & 39.022 & 6.096\\
 & PileBelief & \textbf{6.502} & \textbf{26.244} & 2.883\\
\midrule
1850 & Local & 14.839 & 35.028 & 3.188\\
 & Fixed & 12.065 & 36.175 & \textbf{2.108}\\
 & PileBelief & \textbf{8.320} & \textbf{20.605} & 3.413\\
\bottomrule
\end{tabular}
\end{table}

Here Fixed denotes the first-three-event window, and PileBelief uses the most recent three events. Paired resampling of the six material groups with 10,000 replicates gives descriptive 95\% intervals. Local-minus-PileBelief error differences are 6.048~mm $[2.226,11.289]$ for height, 2.142~L $[-7.549,11.345]$ for bucket volume, and 0.151~kN $[-0.515,0.878]$ for force. Relative to fixed history, the bucket reduction is 13.936~L $[7.864,18.922]$. The intervals capture variation across available material groups, with one training seed and overlapping training sets across folds. Bucket and force differences from Local therefore remain uncertain, whereas height and the bucket advantage over fixed early history have consistent directions in this evaluation.

\subsection{Newly collected target subset}
The 106 newly collected primary targets belong to 17 segments, 15 scene--seed groups, and the 1400 and 1475 material groups. Table~\ref{tab:mpm_added} evaluates the same expanded-data models and uses the same group-then-material aggregation as the main result. Each target's material group remains excluded from its model's training fold. This is a subset of the 164-target evaluation, with no additional training or checkpoint selection. Both groups improve over Local in all three outputs. The added 1475 targets come from one segment, while the remaining targets belong to 1400. The two-group result therefore complements the full six-group result rather than providing balanced coverage of every material.

\begin{table}[!htb]
\caption{MPM errors on the newly collected target subset.}
\label{tab:mpm_added}
\centering\tablesize\setlength{\tabcolsep}{6pt}
\begin{tabular}{@{}lccc@{}}
\toprule
Condition & Height (mm)$\downarrow$ & Bucket (L)$\downarrow$ & Force (kN)$\downarrow$\\
\midrule
Local (empty memory) & 39.361 & 61.872 & 5.236\\
Fixed first three & 30.543 & 72.863 & 5.094\\
\textbf{PileBelief (recent three)} & \textbf{26.321} & \textbf{51.282} & \textbf{4.191}\\
\bottomrule
\end{tabular}
\end{table}

\FloatBarrier
\subsection{Spatial coverage and history masking}
\label{app:mpm_coverage}
We examine the 1400 holdout fold with fixed checkpoints. Every source uses a $5\times5$~m current view, with configured task-region sides of 10 or 15~m. These nominal extents differ from measured pile footprints. Historical extra area is the union of the recent three observation regions outside the current region. Its scene-group mean is 11.37 and 11.25~$\mathrm{m^2}$ in the two 10-m sources and 15.46~$\mathrm{m^2}$ in the 15-m source. Scene extent, geometry, actions, and collection stage vary together, so source differences characterize coverage without isolating an area effect.

Table~\ref{tab:mpm_coverage} uses the same 70 targets from 11 scene--seed groups in the 15-m source. In-view history masks outside-view pixels before memory writes, retaining history order, actions, completed responses, and weights. Unmasked recent history reproduces the primary predictions with the original evaluation batches. Local is the separately trained current-only model from the same fold.

\begin{table}[!htb]
\caption{History masking in the 15-m MPM scenes.}
\label{tab:mpm_coverage}
\centering\tablesize\setlength{\tabcolsep}{6pt}
\begin{tabular}{@{}lccc@{}}
\toprule
Condition & Height (mm)$\downarrow$ & Bucket (L)$\downarrow$ & Force (kN)$\downarrow$\\
\midrule
Local & 69.238 & 108.976 & 5.503\\
In-view history & 42.101 & 87.891 & 5.327\\
\textbf{PileBelief (recent three)} & \textbf{40.821} & \textbf{86.430} & \textbf{5.325}\\
\bottomrule
\end{tabular}
\end{table}

Masking outside-view history increases height error by 1.280~mm and bucket error by 1.461~L, while retaining most of the gain over Local. The retained advantage is consistent with a role for evidence of evolving local conditions in addition to extra spatial coverage. The mask also changes convolution boundaries and write support, making this a fixed-network diagnostic rather than a training-matched branch ablation. The 15-m source occurs only in the held-out 1400 group, so this fold combines material and scene-source shifts. Observation-free MPM rollout and closed-loop action benefits remain outside this single-step evaluation.

\subsection{Simulation and observation resolution}
Delivered configurations share a $0.1$-m MPM grid and 240-Hz simulation frequency. All 337 records with particle-volume fields report approximately $0.7787$~L per reference particle; the manual interruption lacks this field. Height observations use a $33\times25$ grid over $5\times5$~m, with sampling intervals of approximately $0.1563$ and $0.2083$~m. This audit supports consistent recorded resolution across sources, while grid convergence and complete solver equivalence remain untested. Spatial extents are taken from scene configurations and registered coordinates.

\FloatBarrier
\section{Real-data interface and controls}
\label{app:real}
The real-data input contains current pre-excavation geometry and up to three completed interactions. Past crops are encoded in each scoop's local frame and aggregated as temporal context. Planned-action parameters are unavailable to this evaluated interface, which predicts outcomes under the recorded collection policy. The real study therefore evaluates interaction-conditioned terrain and load prediction in its measured sensing configuration.

The train, development, and held-out partitions are defined by continuous segments. The primary errors are segment-equal means, and the score is
\begin{equation}
J_{\mathrm{real}}=\frac12\left(\frac{e_H}{0.1443~\mathrm m}+\frac{e_V}{0.2127~\mathrm{m^3}}\right).
\end{equation}
The reported \method{} means are 112.17~mm and 0.2056~$\mathrm{m^3}$, giving $J_{\mathrm{real}}=0.8720$ under this normalization. The relative joint gain over Local is approximately 4.8\%. Domain-specific scores are interpreted within their own evaluation.

The capacity-matched Local control checks extra network capacity. Zero history zeros standardized evidence while retaining the trained network and its biases. This differs from disabling both correction branches in the reference empty-history path. Matched shuffling substitutes another segment's completed history from the same acquisition date, matching history length and current-crop statistics as closely as possible. These controls distinguish information in correctly matched history from the existence of an additional input branch.

\end{document}